\documentclass{ecai}
\usepackage[utf8]{inputenc}

\usepackage{graphicx}
\usepackage{latexsym}
\usepackage{booktabs}
\usepackage{amsmath,amssymb}

\usepackage{xcolor}
\usepackage{float}

\begin{document}

\begin{frontmatter}

\title{LightSleepX: A Lightweight, Inception-Based Dual-Modal Network for Sleep Staging}
\author{Yi Wang}
\begin{abstract}
Automatic sleep staging is fundamental to personal health monitoring, yet many existing approaches are ill-suited for real-world applications. Traditional pipelines often rely on hand-crafted features or shallow machine learning models that struggle to generalize~\cite{sleepeegnet}, while state-of-the-art deep learning methods—though accurate—are computationally heavy and impractical for resource-constrained environments.

This paper introduces LightSleepX, a lightweight framework designed to deliver robust sleep analysis in resource-constrained environments. LightSleepX integrates EEG and EOG signals through two core design principles optimized for efficiency and adaptability:

\begin{itemize}
    \item \textbf{Efficient Multi-Scale Feature Extraction:} We employ a \textbf{Multi-Branch Inception-style architecture with Depthwise Separable Convolutions} for feature extraction, coupled with \textbf{Multi-scale Enhanced Attention (MEA)} to fuse multi-modal signals. This design captures rich spectral and temporal features with minimal computational overhead.
    
    \item \textbf{Rule-free Temporal Modeling:} We use a \textbf{Mamba encoder} to learn the complex, long-range dependencies between sleep epochs. This allows the model to adaptively understand sleep stage transitions without relying on predefined rules, enhancing its real-world robustness.
\end{itemize}

On public benchmark datasets, LightSleepX demonstrates robust state-of-the-art performance. It achieves 85.9\% accuracy and a 0.803 macro-F1 score on Sleep-EDF-20, and sets a new benchmark on the challenging cross-subject ISRUC-S3 dataset with 81.8\% accuracy and a 0.796 macro-F1 score. Critically, with only 0.049M parameters and 195.9 MFLOPs, our framework is orders of magnitude more efficient, establishing a new benchmark for the trade-off between performance and computational cost. This validates its suitability for practical deployment in everyday health applications where local processing is paramount.
\end{abstract}

\end{frontmatter}


\section{Introduction}

Sleep is a fundamental physiological process crucial for cognitive function, physical health, and overall well-being. It is characterized by distinct stages, including Wake, Non-Rapid Eye Movement (NREM) stages (N1, N2, N3), and Rapid Eye Movement (REM) sleep. Each stage exhibits unique electrophysiological patterns and serves different restorative functions. Sleep staging, the process of classifying these stages from physiological signals, is the cornerstone of sleep medicine, providing critical insights into sleep architecture for diagnosing disorders like insomnia and sleep apnea.

The clinical gold standard for this analysis is polysomnography (PSG), where multiple physiological signals are recorded overnight. Trained experts then manually score the data in 30-second epochs according to established guidelines~\cite{Iber2007, Berry2012}. However, this process is cumbersome, expensive, and reliant on manual scoring, making it unsuitable for continuous, real-world tracking in home environments. This limitation has motivated the development of automated systems for sleep analysis.

Recent advances in deep learning have enabled data-driven methods that learn powerful representations directly from raw EEG and EOG signals. Examples include DeepSleepNet~\cite{deepsleepnet}, SleepEEGNet~\cite{sleepeegnet}, AttnSleep~\cite{attnsleep}, U-Time~\cite{utime}, and the Sleep Transformer~\cite{sleeptransformer}, all achieving state-of-the-art accuracy. However, these models often employ large and complex architectures with millions of parameters, resulting in high computational and energy demands that limit deployment on resource-constrained devices. In addition, privacy concerns are paramount in home monitoring scenarios, where local on-device processing is preferred to avoid transmitting sensitive physiological data. Conversely, lightweight methods relying on hand-crafted features or shallow classifiers fail to capture the rich temporal dynamics of sleep, resulting in sub-optimal performance under real-world conditions.

To address these challenges, we propose \textbf{LightSleepX}, a dual-modal framework designed for efficient sleep staging that balances accuracy, efficiency, and privacy. Our approach leverages two key principles:
\begin{enumerate}
  \item \textbf{Efficient Multi-Scale Feature Extraction}, using a multi-branch Inception-style backbone with depthwise separable convolutions and \textbf{Multi-scale Enhanced Attention (MEA)} for multi-modal fusion, and
  \item \textbf{Rule-free Temporal Modeling}, employing a \textbf{Mamba encoder} to capture long-range dependencies without relying on predefined transition rules.
\end{enumerate}

Our main contributions are as follows:
\begin{itemize}
    \item \textbf{A Practical Framework for Local Sleep Staging:} We present \textbf{LightSleepX}, a lightweight framework engineered for efficient local data processing, enabling privacy-preserving, longitudinal sleep tracking in natural home environments.
    \item \textbf{An Efficient and Adaptive Architecture:} We combine a multi-branch Inception-style feature extractor (using depthwise separable convolutions) with a Mamba encoder for efficient feature learning and rule-free temporal modeling, significantly reducing parameter count and computational cost.
    \item \textbf{State-of-the-Art Performance with Minimal Footprint:} LightSleepX achieves competitive accuracy (e.g., 85.9\% on public benchmarks) and macro-F1 scores with only 0.049M parameters, demonstrating its suitability for resource-limited, privacy-sensitive deployment.
\end{itemize}

\section{Methodology}

\subsection{Overall Architecture}

\begin{figure*}[t!]
    \centering
    \includegraphics[width=\textwidth]{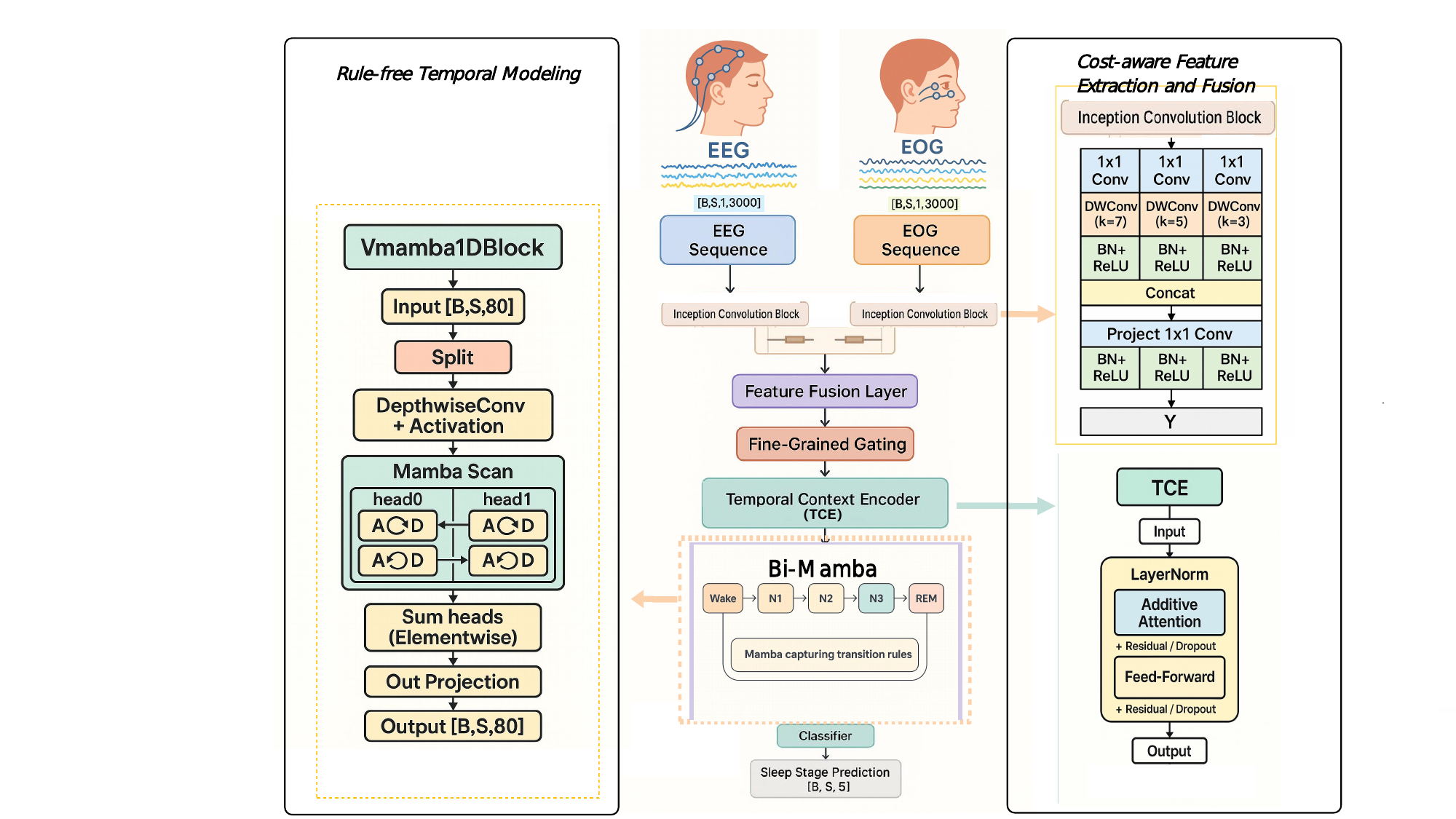}
    \caption{The overall architecture of LightSleepX. A sequence of 30-second epochs from EEG and EOG signals are fed into parallel feature extractors. The extracted features are then fused by the MEA module. A bidirectional Mamba encoder models the temporal dependencies across the sequence of epochs, and a final MLP head classifies the center epoch of the sequence.}
    \label{fig:overall_architecture}
\end{figure*}

The architecture of LightSleepX is meticulously designed for a harmonious balance between high accuracy and computational efficiency, making it ideal for deployment on resource-constrained devices. As illustrated in Figure~\ref{fig:overall_architecture}, the model processes sequences of raw EEG and EOG signals through four primary stages: (1) an efficient multi-scale feature extractor, (2) a multi-modal fusion block, (3) a rule-free temporal encoder, and (4) a final classification head.

The data processing pipeline is as follows:
\begin{enumerate}
    \item \textbf{Input and Preprocessing}: The input to the model is a sequence of $S$ consecutive 30-second epochs of raw EEG and EOG signals. Each epoch is normalized using z-score. The input tensor for each modality has a shape of $(B, S, 1, T)$, where $B$ is the batch size, $S$ is the sequence length (e.g., 10), and $T$ is the number of samples per epoch (e.g., 3000 for 100 Hz).

    \item \textbf{Multi-Scale Feature Extractor}: Each epoch within the sequence is processed independently by our \textbf{Multi-Branch Inception-style Feature Extractor}. This module, detailed in Section~\ref{sec:feature_extractor}, employs parallel branches of depthwise separable convolutions with varying kernel sizes to efficiently capture features at multiple temporal resolutions from the raw signal.

    \item \textbf{Multi-Modal Fusion}: The feature maps extracted from the EEG and EOG modalities are concatenated and then enhanced by the \textbf{Multi-scale Enhanced Attention (MEA)} module. As described in Section~\ref{sec:fusion}, MEA adaptively recalibrates channel-wise feature responses, promoting salient information for robust fusion.

    \item \textbf{Temporal Encoder}: The sequence of fused feature vectors, representing consecutive epochs, is fed into a \textbf{bidirectional Mamba encoder}. This powerful State Space Model, detailed in Section~\ref{sec:temporal_encoder}, captures the complex, long-range dependencies and transition rules between sleep stages.

    \item \textbf{Classification Head}: Finally, the context-aware representation for each epoch from the Mamba encoder is passed to a simple Multilayer Perceptron (MLP) head, which outputs the probability distribution over the five sleep stages (Wake, N1, N2, N3, REM).
\end{enumerate}

\subsection{Key Architectural Modules}

This section provides a detailed breakdown of the LightSleepX framework, strictly adhering to the model configuration defined in our experiments. Each component and its key parameters are described to ensure full reproducibility.

\subsubsection{Datasets and Preprocessing}

\paragraph{Sleep-EDF Expanded.} We use two standard subsets of the public Sleep-EDF Expanded dataset. \textbf{Sleep-EDF-20} contains 42,308 sleep epochs from 20 healthy subjects (10 male, 10 female, aged 25-34). Each subject has two full-night PSG recordings, except for one subject with a single recording due to device failure. \textbf{Sleep-EDF-78} is a larger and more diverse subset with 195,479 epochs from 78 healthy subjects (aged 25-101). For both datasets, we follow standard protocols, using the Fpz-Cz EEG and ROC-LOC EOG channels, both sampled at 100 Hz.

\paragraph{ISRUC-Sleep.} We also evaluate on the challenging \textbf{ISRUC-Sleep (Subgroup 3)} dataset, which includes 100 subjects~\cite{dan}. This dataset is known for its high inter-subject variability, making it an excellent benchmark for model generalization. We follow the standard 10-fold cross-validation protocol, where data from 80\% of subjects is used for training, 10\% for validation, and 10\% for testing in each fold.

\paragraph{Preprocessing.} For all datasets, we apply a consistent preprocessing pipeline for fair comparison. Raw signals are segmented into 30-second epochs. Following the AASM standard, we merge the N3 and N4 stages into a single N3 stage. Each epoch is then independently normalized using a z-score to mitigate signal amplitude variations across subjects and recordings.

\paragraph{Sequential Input Construction.}
To model temporal dependencies, we construct input sequences of length \(S=10\) using a sliding window with a stride of 1. This results in each input sample to the model representing a 5-minute segment of sleep data.

\subsubsection{Multi-Branch Inception-Style Feature Extractor}
\label{sec:feature_extractor}

Our feature extraction stage is designed for efficiency, adopting the principles of modern, efficient convolutional architectures. To capture a rich hierarchy of features while maintaining computational efficiency, we introduce a \textbf{Multi-Branch Inception-style architecture}. This design draws inspiration from the multi-scale philosophy of Inception networks~\cite{Szegedy2016Rethinking}. The module is composed of multiple parallel branches, each configured with a different kernel size (e.g., 7, 5, and 3). This allows the model to simultaneously process the input signal at different resolutions, capturing both fine-grained local patterns and broader contextual features.

This multi-scale approach is particularly well-suited for sleep signal analysis. Sleep EEG is characterized by a diverse set of transient and oscillatory events with varying durations, such as slow, high-amplitude delta waves (dominant in N3 sleep), and short, high-frequency sleep spindles (a hallmark of N2 sleep). By using parallel branches with different kernel sizes, our model can concurrently capture these distinct patterns: larger kernels are adept at identifying long-duration slow-wave activity, while smaller kernels can precisely detect brief events like spindles.

Crucially, to maintain a low parameter count, each branch utilizes \textbf{depthwise separable convolutions}~\cite{Howard2017MobileNets}. The operation for a single branch with kernel size $k$ can be formulated as:
\begin{equation}
    \text{Branch}_k(\mathbf{x}) = \text{PointConv}(\text{DepthConv}_k(\mathbf{x}))
\end{equation}
where $\mathbf{x}$ is the input feature map, $\text{DepthConv}_k$ is a depthwise convolution with kernel size $k$, and $\text{PointConv}$ is a pointwise (1x1) convolution.

The outputs from all parallel branches are then concatenated and fused via a final pointwise convolution to produce the final feature map $\mathbf{y}$:
\begin{equation}
    \mathbf{y} = \text{PointConv}(\text{Concat}[\text{Branch}_{k_1}(\mathbf{x}), \dots, \text{Branch}_{k_N}(\mathbf{x})])
\end{equation}
This strategy effectively combines the multi-scale feature extraction capabilities of Inception with the efficiency of depthwise separable convolutions. For our specific implementation, the module uses three branches with kernel sizes \{3, 5, 7\} and produces a fused feature map with 128 channels for each modality.

\subsubsection{Multi-Modal Fusion and Enhancement}
\label{sec:fusion}

The features extracted from the EEG and EOG streams are first fused via simple \textbf{concatenation} along the channel dimension. This operation preserves the distinct information from each modality. The resulting combined feature map, $\mathbf{F} \in \mathbb{R}^{C \times T}$, is then processed by a \textbf{Multi-scale Enhanced Attention (MEA)} module.

Inspired by the EMA method~\cite{ouyang2023ema}, MEA functions as an efficient channel-wise attention mechanism that captures dependencies across multiple scales. It consists of two parallel branches. One branch uses a 1D convolution with a small kernel size (e.g., 3) to capture local cross-channel interactions, while the other branch uses a larger kernel (e.g., 5) to model more global interactions. The attention weights from both branches are aggregated and applied to the input features.

The process can be summarized as:
\begin{align}
    \mathbf{a}_1 &= \text{Conv1D}_{k_1}(\text{GAP}(\mathbf{F})) \\
    \mathbf{a}_2 &= \text{Conv1D}_{k_2}(\text{GAP}(\mathbf{F})) \\
    \mathbf{A} &= \sigma(\mathbf{a}_1 + \mathbf{a}_2)
\end{align}
where $\text{GAP}$ is Global Average Pooling across the temporal dimension, $\text{Conv1D}_{k}$ is a 1D convolution with kernel size $k$, and $\sigma$ is the sigmoid activation function. The final enhanced feature map $\mathbf{F'}$ is obtained by re-weighting the input features:
\begin{equation}
    \mathbf{F'} = \mathbf{F} \odot \mathbf{A}
\end{equation}
This allows the model to dynamically emphasize the most salient modality (EEG or EOG) or specific features for a given epoch, enabling effective and adaptive fusion.

\subsubsection{Rule-Free Temporal Modeling with Mamba}
\label{sec:temporal_encoder}

Our framework uses a data-driven approach to model the complex temporal dynamics of sleep, moving beyond manually defined rules. The core of our temporal modeling is the \textbf{Mamba encoder}~\cite{mamba2023}, a type of State Space Model (SSM) designed for efficient long-range dependency capturing.

At its core, a continuous SSM maps a 1-D input signal $u(t)$ to an output $y(t)$ through a latent state $h(t)$:
\begin{align}
    h'(t) &= \mathbf{A}h(t) + \mathbf{B}u(t) \\
    y(t) &= \mathbf{C}h(t) + \mathbf{D}u(t)
\end{align}
where $\mathbf{A} \in \mathbb{R}^{N \times N}$ is the state transition matrix, and $\mathbf{B} \in \mathbb{R}^{N \times 1}$, $\mathbf{C} \in \mathbb{R}^{1 \times N}$ are projection matrices. To be used in a deep learning model, this system is discretized using a timestep $\Delta$. Mamba employs a zero-order hold (ZOH) to transform the continuous parameters $(\mathbf{A}, \mathbf{B})$ to discrete parameters $(\bar{\mathbf{A}}, \bar{\mathbf{B}})$:
\begin{align}
    \bar{\mathbf{A}} &= \exp(\Delta \mathbf{A}) \\
    \bar{\mathbf{B}} &= (\exp(\Delta \mathbf{A}) - \mathbf{I})\mathbf{A}^{-1}\mathbf{B}
\end{align}
The discrete SSM is then computed as $h_k = \bar{\mathbf{A}}h_{k-1} + \bar{\mathbf{B}}u_k$ and $y_k = \mathbf{C}h_k + \mathbf{D}u_k$.

Mamba's key innovation is its \textbf{selection mechanism}, which makes the SSM parameters data-dependent. Specifically, the timestep $\Delta$ and the projection matrices $\mathbf{B}$ and $\mathbf{C}$ are derived from the input sequence itself, allowing the model to selectively focus on or ignore information at each time step.

Our implementation, the \texttt{Vmamba1DBlock}, follows this selective SSM principle. For an input sequence $\mathbf{x} \in \mathbb{R}^{L \times D}$, the block performs the following steps:
\begin{enumerate}
    \item \textbf{Linear Projection \& Gating}: The input $\mathbf{x}$ is projected to create two intermediate representations, $\mathbf{x}_{\text{conv}}$ and a gating signal $\mathbf{z}$.
    \begin{equation}
        \mathbf{x}_{\text{conv}}, \mathbf{z} = \text{Split}(\text{Linear}(\mathbf{x}))
    \end{equation}
    \item \textbf{1D Causal Convolution}: A 1D causal depthwise convolution is applied to $\mathbf{x}_{\text{conv}}$, followed by a SiLU activation to produce the main input $\mathbf{u}$ for the SSM.
    \begin{equation}
        \mathbf{u} = \text{SiLU}(\text{Conv1D}(\mathbf{x}_{\text{conv}}))
    \end{equation}
    \item \textbf{Selective Scan}: The core SSM operation is performed on $\mathbf{u}$ using \texttt{selective\_scan\_fn}, which implements the efficient parallel scan algorithm to compute the output sequence $\mathbf{y}_{\text{scan}}$.
    \begin{equation}
        \mathbf{y}_{\text{scan}} = \text{SSM}(\Delta, \mathbf{A}, \mathbf{B}, \mathbf{C})(\mathbf{u})
    \end{equation}
    \item \textbf{Gated Output}: The output of the scan is modulated by the gating signal $\mathbf{z}$ and projected back to the original dimension.
    \begin{equation}
        \mathbf{y}_{\text{out}} = \text{Linear}(\mathbf{y}_{\text{scan}} \odot \mathbf{z})
    \end{equation}
\end{enumerate}

To leverage both past and future context for sleep staging, we employ a \textbf{bidirectional} Mamba architecture. This is achieved by processing the input sequence with two independent Mamba blocks: one in the forward direction and another in the backward (reversed sequence) direction. The final output is the sum of the outputs from both blocks. This standard strategy allows the model to build a complete contextual understanding for each sleep epoch while preserving the computational efficiency of the underlying causal scan mechanism.

\subsubsection{Classification Head}

\paragraph{MLP Classifier.}
The final classification is performed by a lightweight \textbf{multilayer perceptron (MLP)}. It takes the context-aware feature vector for each epoch from the Mamba encoder and maps it to the final logits for the five sleep stages (Wake, N1, N2, N3, REM). Its simplicity ensures that the classification step does not become a computational bottleneck, preserving the overall efficiency of the framework.

\subsection{Experimental Setup}

Our model is trained using a carefully designed strategy to ensure optimal performance and generalization capability:

\begin{itemize}
    \item \textbf{Batch Size}: 128 samples per batch, balancing computational efficiency and gradient stability.
    
    \item \textbf{Optimizer}: Adam optimizer with learning rate of 0.001, weight decay of 0.001, and amsgrad=true to address potential issues with adaptive learning rate methods.
    
    \item \textbf{Loss Function}: We use a \textbf{Dynamic Focal Loss} to address the significant class imbalance inherent in sleep data. Focal Loss adapts the standard cross-entropy loss by down-weighting the contribution of well-classified examples, thereby focusing the model's training on more challenging samples~\cite{ref7}. The loss is defined as:
    \begin{equation}
    \mathcal{L}_{FL} = -(1 - \hat{y}_i)^\gamma \log(\hat{y}_i)
    \end{equation}
    where $\hat{y}_i$ is the predicted probability for the true class, and $\gamma$ is the focusing parameter (set to 2.0 in our experiments), which modulates the down-weighting effect. This allows the model to effectively learn from minority classes like N1 and N3.
    
    \item \textbf{Training Duration}: The model is trained for 100 epochs, with early stopping based on validation performance to prevent overfitting.
    
    \item \textbf{Evaluation Metrics}: Performance is evaluated using accuracy and macro F1-score across all five sleep stages (Wake, N1, N2, N3, and REM). The macro F1-score is particularly important due to the class imbalance in sleep data, providing a balanced measure of performance across all stages:
    \begin{equation}
    \text{F1}_{\text{macro}} = \frac{1}{C}\sum_{i=1}^{C} \text{F1}_i
    \end{equation}
    where $\text{F1}_i$ is the F1-score for class $i$.
\end{itemize}

The model is implemented in PyTorch and trained on NVIDIA GPUs. K-fold cross-validation with subject-based splits ensures robust evaluation and prevents data leakage between training and testing sets.

\section{Results and Discussion}

\subsection{Comparison with State-of-the-Art Methods}
We evaluated LightSleepX against a comprehensive set of state-of-the-art methods on the Sleep-EDF-20, Sleep-EDF-78, and ISRUC-S3 datasets. The results, summarized in Tables~\ref{tab:edf20_comparison}, \ref{tab:edf78_comparison}, and \ref{tab:isruc_comparison}, demonstrate our model's robust and superior performance, particularly on metrics crucial for imbalanced data.

\begin{table}[H]
\centering
\caption{Performance comparison on \textbf{Sleep-EDF-20} against state-of-the-art models including SleepEEGNet~\cite{sleepeegnet}, DeepSleepNet~\cite{deepsleepnet}, AttnSleep~\cite{attnsleep}, SeqSleepNet~\cite{seqsleepnet}, and TinySleepNet~\cite{tinysleepnet}. Best results are in \textbf{bold}.}
\label{tab:edf20_comparison}
\begin{tabular}{lcccc}
\toprule
\textbf{Method} & \textbf{\#Param (M)} & \textbf{ACC (\%)} & \textbf{Macro-F1} & \textbf{$\kappa$} \\
\midrule
SleepEEGNet~\cite{sleepeegnet}   & 2.6 & 81.5 & 76.6 & 0.75 \\
DeepSleepNet~\cite{deepsleepnet}  & 21 & 82.0 & 76.9 & 0.76 \\
IITNet\cite{iitnet} & - & 83.9 & 77.6 & 0.78 \\
AttnSleep~\cite{attnsleep}     & 0.6 & 84.4 & 78.1 & 0.79 \\
SeqSleepNet~\cite{seqsleepnet}   & 0.164 & 84.6 & 78.0 & 0.79 \\
TinySleepNet~\cite{tinysleepnet}  & 1.3 & 85.4 & 80.5 & 0.80 \\
\midrule
\textbf{LightSleepX (Ours)} & \textbf{0.049} & \textbf{85.9} & \textbf{80.7} & \textbf{0.81} \\
\bottomrule
\end{tabular}
\end{table}

\begin{table}[H]
\centering
\small
\caption{Performance and efficiency comparison on \textbf{Sleep-EDF-78} against state-of-the-art models including SleepEEGNet~\cite{sleepeegnet}, U-Time~\cite{utime}, SleepTransformer~\cite{sleeptransformer}, AttnSleep~\cite{attnsleep}, SeqSleepNet~\cite{seqsleepnet}, and TinySleepNet~\cite{tinysleepnet}. Best results are in \textbf{bold}.}
\label{tab:edf78_comparison}
\resizebox{\columnwidth}{!}{%
\begin{tabular}{lcccc}
\toprule
\textbf{Method} & \textbf{\#Param (M)} & \textbf{ACC (\%)} & \textbf{Macro-F1} & \textbf{$\kappa$} \\
\midrule
SleepEEGNet~\cite{sleepeegnet}       & 2.6  & 80.0 & 73.6 & 0.73 \\
U-Time~\cite{utime}            & 1.1  & 81.3 & 76.3 & 0.75 \\
SleepTransformer~\cite{sleeptransformer}  & 3.7  & 81.4 & 74.3 & 0.74 \\
AttnSleep~\cite{attnsleep}         & 5.2  & 81.3 & 75.1 & 0.74 \\
SeqSleepNet~\cite{seqsleepnet}       & 0.164& 82.6 & 76.3 & 0.76 \\
TinySleepNet~\cite{tinysleepnet}      & 1.3  & \textbf{83.1} & 78.1 & \textbf{0.77} \\
\midrule
\textbf{LightSleepX (Ours)} & \textbf{0.049} & 82.6 & \textbf{78.7} & \textbf{0.77} \\
\bottomrule
\end{tabular}%
}
\end{table}

\begin{table}[H]
\centering
\caption{Performance comparison on \textbf{ISRUC-S3 (10-fold)} against state-of-the-art models including DeepSleepNet~\cite{deepsleepnet}, AttnSleep~\cite{attnsleep}, and DAN~\cite{dan}. Best results are in \textbf{bold}.}
\label{tab:isruc_comparison}
\begin{tabular}{lcccc}
\toprule
\textbf{Method} & \textbf{\#Param (M)} & \textbf{ACC (\%)} & \textbf{Macro-F1} & \textbf{$\kappa$} \\
\midrule
DeepSleepNet~\cite{deepsleepnet}             & 21 & 74.56 & 0.739 & 0.676 \\
AttnSleep~\cite{attnsleep}                & 0.6 & 76.11 & 0.740 & 0.693 \\
DAN~\cite{dan}                      & - & 76.87 & 0.744 & 0.701 \\
\midrule
\textbf{LightSleepX (Ours)} & \textbf{0.049} & \textbf{81.80} & \textbf{0.796} & \textbf{0.761} \\
\bottomrule
\end{tabular}
\end{table}

\subsection{Performance Analysis}
\begin{figure*}[t!]
    \centering
    \includegraphics[width=0.3\textwidth]{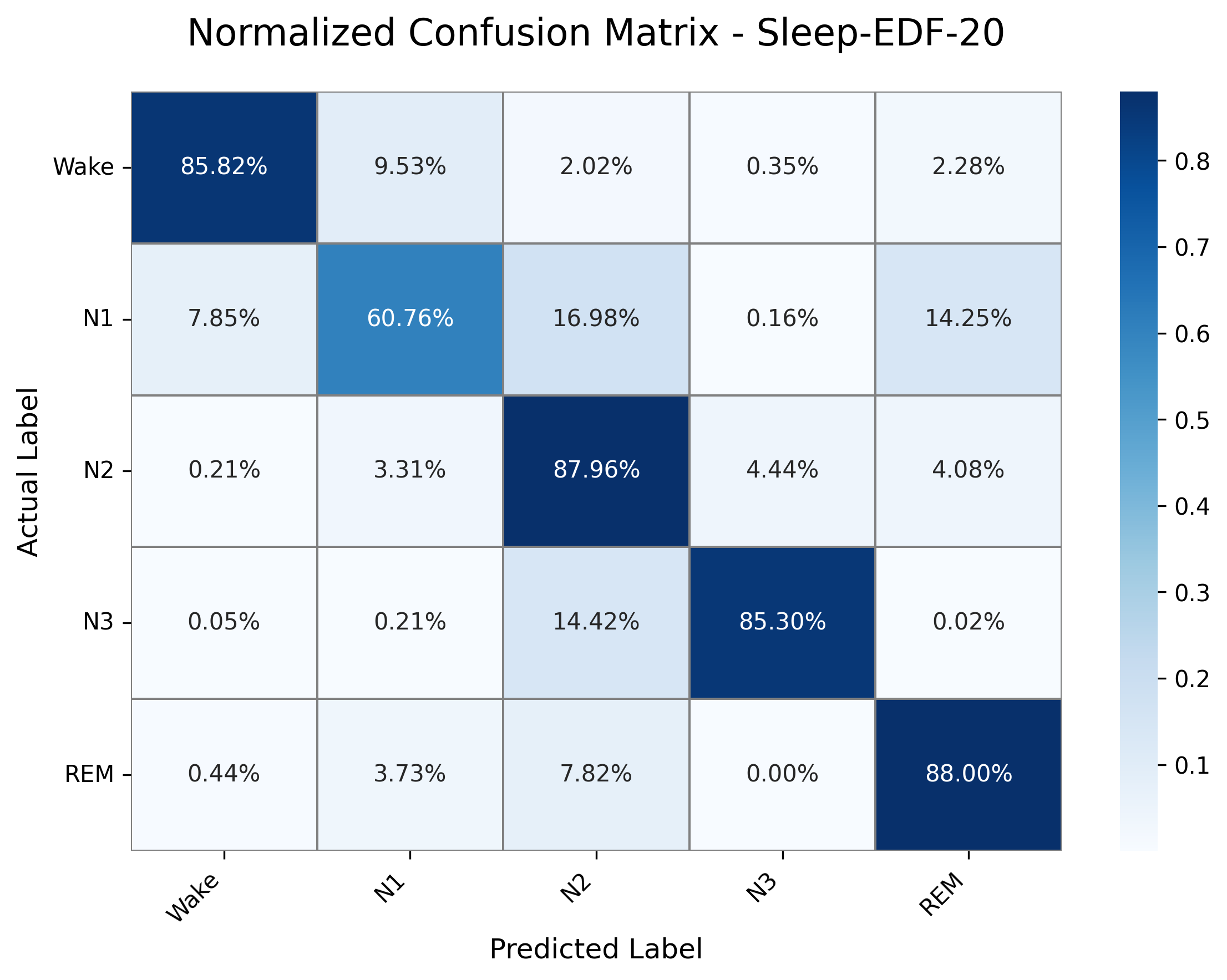}
    \hfill
    \includegraphics[width=0.3\textwidth]{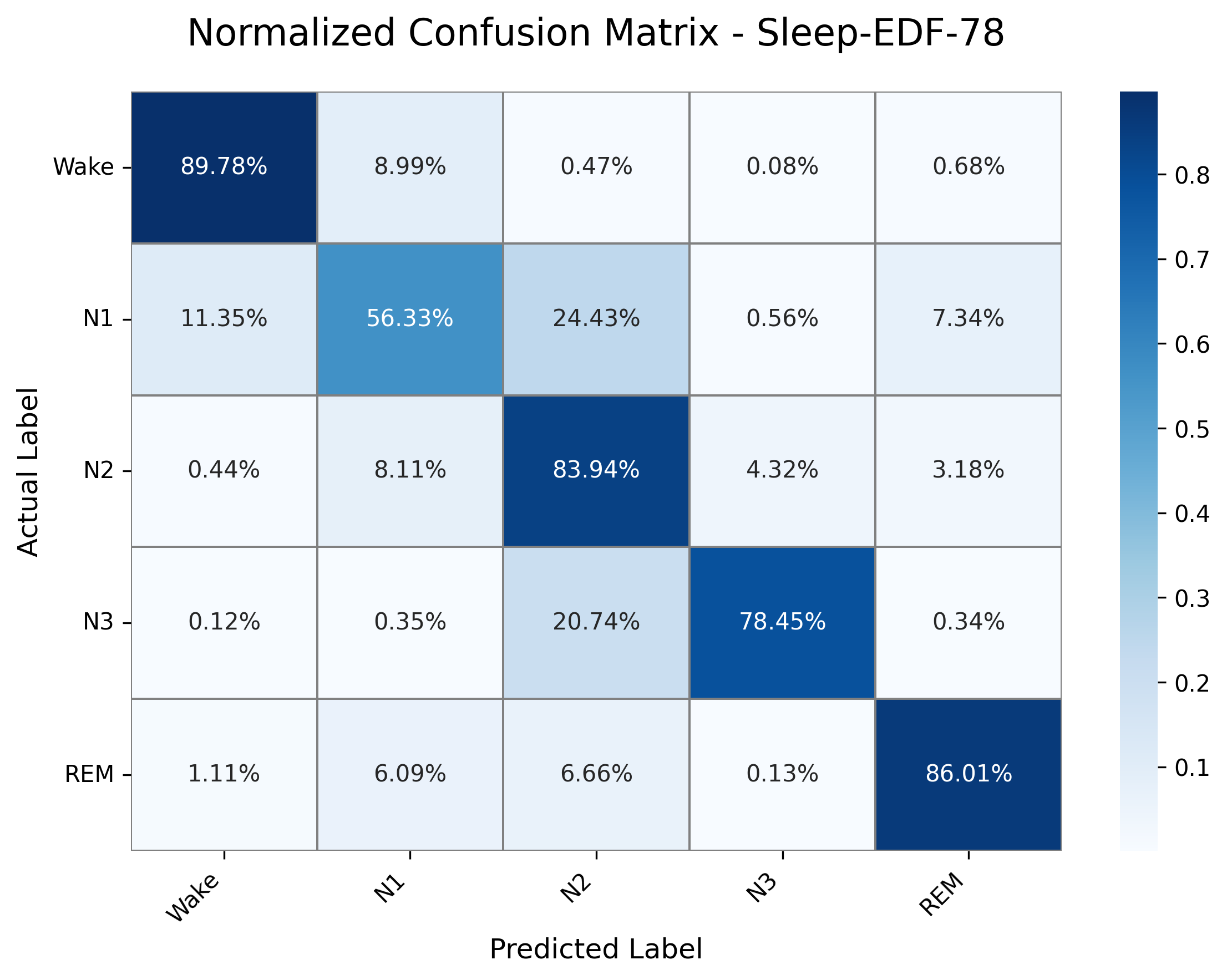}
    \hfill
    \includegraphics[width=0.3\textwidth]{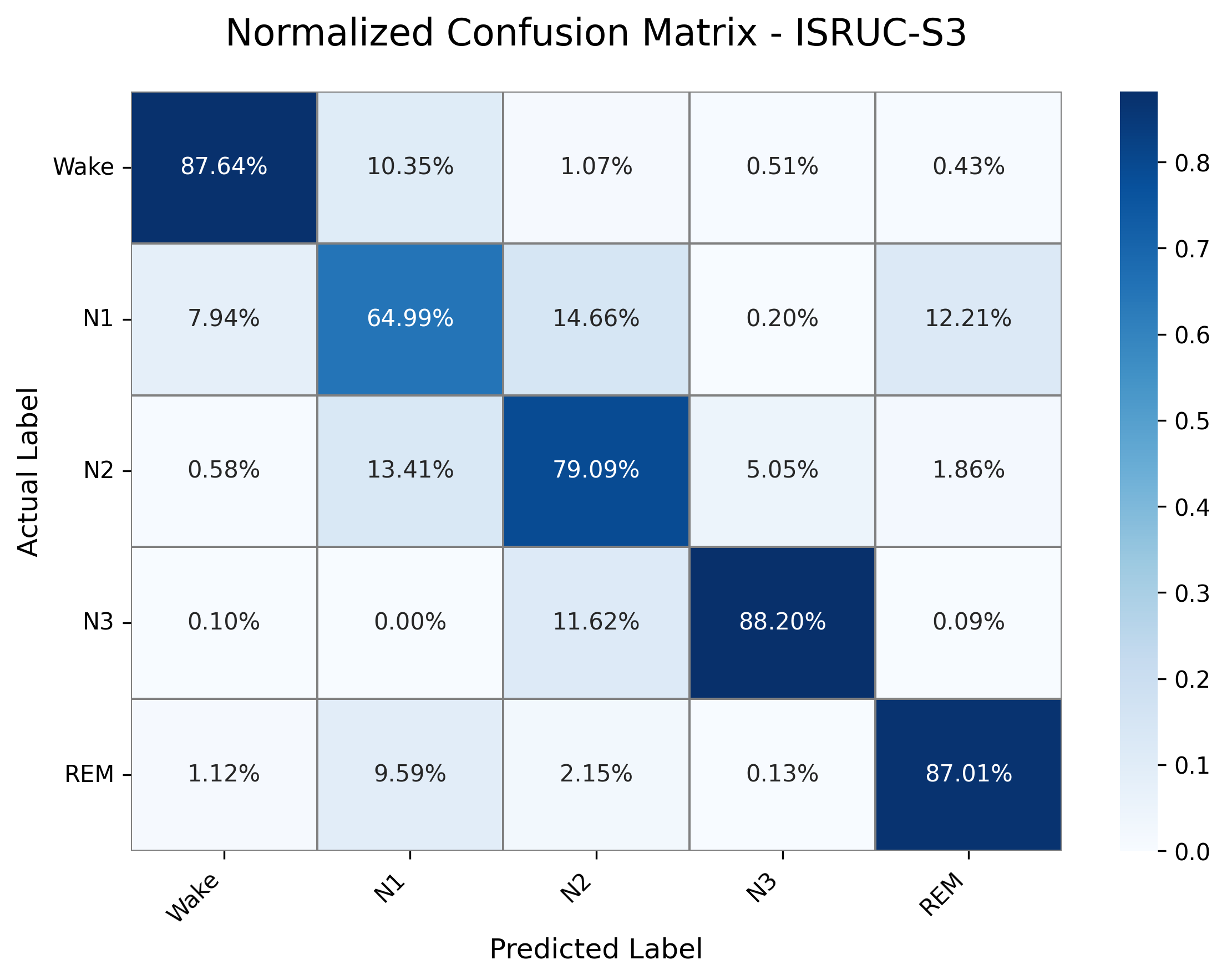}
    \caption{Normalized confusion matrices on the Sleep-EDF-20, Sleep-EDF-78, and ISRUC-S3 datasets. The diagonal elements represent the percentage of correctly classified epochs for each sleep stage, while off-diagonal elements indicate misclassifications. These visualizations offer a granular view of the model's performance across different sleep stages.}
    \label{fig:confusion_matrices}
\end{figure*}

LightSleepX establishes a new state-of-the-art in robust and efficient sleep staging.
On \textbf{Sleep-EDF-20} (Table \ref{tab:edf20_comparison}), it achieves top performance across all metrics (85.9\% ACC, 80.7\% Macro-F1, 0.81 $\kappa$).
On the more challenging \textbf{Sleep-EDF-78} (Table \ref{tab:edf78_comparison}), it obtains the highest Macro-F1 score (78.7\%), demonstrating superior handling of class imbalance.
Most notably, on the cross-subject \textbf{ISRUC-S3} dataset (Table~\ref{tab:isruc_comparison}), LightSleepX sets a new benchmark with 81.80\% accuracy and a 0.796 Macro-F1, a significant leap of over 5 percentage points in accuracy compared to previous methods, validating its generalization capability. The normalized confusion matrices in Figure~\ref{fig:confusion_matrices} provide a more detailed breakdown of the classification performance across all stages for each dataset.

A key advantage is its strong performance on the notoriously difficult N1 stage, where it achieves a state-of-the-art F1-score of 63.0\% (Table~\ref{tab:n1_comparison}).

\begin{table}[H]
\centering
\caption{N1 F1-Score comparison on \textbf{Sleep-EDF-78} against models like AttnSleep~\cite{attnsleep}, SeqSleepNet~\cite{seqsleepnet}, TinySleepNet~\cite{tinysleepnet}, and U-Time~\cite{utime}. Best result is in \textbf{bold}.}
\label{tab:n1_comparison}
\begin{tabular}{lc}
\toprule
\textbf{Method} & \textbf{N1 F1-Score (\%)} \\
\midrule
AttnSleep~\cite{attnsleep} & 42.0 \\
SeqSleepNet~\cite{seqsleepnet} & 46.0 \\
TinySleepNet~\cite{tinysleepnet} & 51.0 \\
U-Time~\cite{utime} & 51.0 \\
\midrule
\textbf{LightSleepX (Ours)} & \textbf{63.0} \\
\bottomrule
\end{tabular}
\end{table}

Crucially, this state-of-the-art performance is achieved within an exceptionally lightweight framework. As shown in Table~\ref{tab:edf78_comparison}, with only \textbf{0.049M parameters}, our model is approximately \textbf{3.3 times smaller} than the next-lightest model, SeqSleepNet (0.164M), and over \textbf{22 times smaller} than competitors like U-Time (1.1M) and TinySleepNet (1.3M). This remarkable efficiency, combined with top-tier performance, establishes a new benchmark for the trade-off between accuracy and computational cost.

\section{Conclusion}

In this paper, we introduced \textbf{LightSleepX}, a dual-modal deep learning framework engineered for efficient and accurate sleep staging. By combining a \textbf{Multi-Branch Inception-style} feature extractor with a \textbf{bidirectional Mamba encoder}, our model establishes a new state-of-the-art in the balance between performance and computational cost.
Experimental results show that LightSleepX achieves state-of-the-art performance, particularly on Macro-F1 scores, while using only \textbf{0.049M parameters}. Its high accuracy, exceptional handling of class imbalance (especially for the N1 stage), and minimal footprint make it a powerful, practical, and privacy-preserving solution for the next generation of local sleep monitoring systems.

Future work will focus on three key directions: (1) expanding the framework to incorporate additional physiological signals (e.g., EMG, ECG) to enhance clinical relevance; (2) pursuing hardware-aware optimizations, such as quantization and pruning, for deployment on edge-computing platforms; and (3) investigating semi-supervised and domain-adaptation techniques to improve generalization across diverse populations.

\subsection{Ablation Study}
\label{sec:ablation_study}

To quantify the contribution of each key component in the proposed LightSleepX framework, we conducted a set of ablation experiments on the Sleep-EDF20 dataset. Specifically, we systematically removed or replaced the following modules:
\begin{itemize}
    \item \textbf{Multi-Branch Inception-style Feature Extractor}: replaced with a standard CNN backbone.
    \item \textbf{Multi-scale Enhanced Attention (MEA)}: removed the MEA module for multi-modal fusion.
    \item \textbf{Mamba Encoder}: replaced with a Bi-LSTM encoder for temporal modeling.
\end{itemize}

We evaluated each model variant using Accuracy (ACC), macro-averaged F1-score (Macro-F1), and the total number of parameters. The results are summarized in Table~\ref{tab:ablation_study}.

\begin{table}[ht]
\centering
\caption{Ablation study on the Sleep-EDF20 dataset. Each row removes or substitutes a key module to quantify its contribution.}
\label{tab:ablation_study}
\begin{tabular}{lcc}
\toprule
\textbf{Model Variant} & \textbf{ACC (\%)} & \textbf{Macro-F1} \\
\midrule
Full Model (LightSleepX) & \textbf{85.9} & \textbf{0.803} \\
w/o Multi-Branch Inception-style (Std. CNN)  & 80.5 & 0.750 \\
w/o MEA                  & 81.8 & 0.755 \\
w/o Mamba (Bi-LSTM)      & 81.2 & 0.748 \\
\bottomrule
\end{tabular}
\end{table}

From Table~\ref{tab:ablation_study}, we make the following observations:

\begin{itemize}
    \item \textbf{Multi-Branch Inception-style Feature Extractor}: Replacing the Inception-style block with a standard CNN causes a notable drop in performance (e.g., accuracy decreases by 5.4\%), highlighting the importance of efficient multi-scale feature learning for EEG/EOG signals.
    \item \textbf{Multi-scale Enhanced Attention}: Removing MEA results in a decline in both ACC and Macro-F1, demonstrating its critical role in adaptive multi-modal fusion.
    \item \textbf{Mamba Encoder}: Substituting the Mamba encoder with Bi-LSTM degrades model performance, indicating the superiority of Mamba for capturing long-range temporal dependencies efficiently.
\end{itemize}

Overall, the ablation study demonstrates that each proposed module contributes significantly to the overall performance and efficiency of LightSleepX, confirming the effectiveness of our architectural choices.

\end{document}